\documentclass[11pt]{article} %
\usepackage{rldmsubmit,palatino}
\usepackage{graphicx}

\usepackage{subcaption} %

\usepackage{hyperref}
\usepackage{float}
\usepackage{amsfonts}
\usepackage[%
backend=biber,%
natbib=true,%
style=apa,%
uniquename=false,%
useprefix=false%
]{biblatex}
\title{Upside Down Reinforcement Learning with Policy Generators}

\author{
Jacopo Di Ventura \thanks{Now at Leiden University. Correspondence to  \href{mailto:j.di.ventura@liacs.leidenuniv.nl}{\texttt{j.di.ventura@liacs.leidenuniv.nl}}}\enspace$^{1}$ \\
\And
Dylan R.~Ashley $^{1,2,3,4}$ \\
\And
Vincent Herrmann $^{1,2,3}$ \\
\And
Francesco Faccio $^{1,2,3,4}$ \\
\And
J{\"{u}}rgen Schmidhuber $^{1,2,3,4,5}$ \\
\AND
{\normalfont $^1$ Universit{\`{a}} della Svizzera italiana, Lugano, Switzerland} \\
$^2$ The Swiss AI Lab IDSIA (USI-SUPSI), Lugano, Switzerland \\
$^3$ Scuola universitaria professionale della Svizzera italiana, Lugano, Switzerland \\
$^4$ Center of Excellence for Generative AI, King Abdullah University of Science and Technology, Thuwal, Saudi Arabia \\
$^5$ NNAISENSE, Lugano, Switzerland \\
}

\begin{document}

\maketitle

\begin{abstract}

Upside Down Reinforcement Learning (UDRL) is a promising framework for solving reinforcement learning problems which focuses on learning command-conditioned policies.
In this work, we extend UDRL to the task of learning a command-conditioned generator of deep neural network policies.
We accomplish this using Hypernetworks---a variant of Fast Weight Programmers, which learn to decode input commands representing a desired expected return into command-specific weight matrices.
Our method, dubbed Upside Down Reinforcement Learning with Policy Generators (UDRLPG), streamlines comparable techniques by removing the need for an evaluator or critic to update the weights of the generator.
To counteract the increased variance in last returns caused by not having an evaluator, we decouple the sampling probability of the buffer from the absolute number of policies in it, which, together with a simple weighting strategy, improves the empirical convergence of the algorithm.
Compared with existing algorithms, UDRLPG achieves competitive performance and high returns, sometimes outperforming more complex architectures.
Our experiments show that a trained generator can generalize to create policies that achieve unseen returns zero-shot.
The proposed method appears to be effective in mitigating some of the challenges associated with learning highly multimodal functions.
Altogether, we believe that UDRLPG represents a promising step forward in achieving greater empirical sample efficiency in RL.
A full implementation of UDRLPG is publicly available at \href{https://github.com/JacopoD/udrlpg_}{\texttt{https://github.com/JacopoD/udrlpg\_}}

\end{abstract}


\acknowledgements{This work was supported by the European Research Council (ERC, Advanced Grant Number 742870) and the Center of Excellence for Generative AI at the King Abdullah University of Science and Technology (KAUST, Award Number 5940). We also want to thank both the NVIDIA Corporation for donating a DGX-1 as part of the Pioneers of AI Research Award and IBM for donating a Minsky machine.}

\startmain %

\section{Introduction}

Reinforcement Learning (RL) is a powerful framework for solving sequential decision-making problems. In RL, the standard approach to policy optimization typically involves training a policy network to maximize expected returns. Upside Down RL~\citep{schmidhuber2019reinforcement}, or UDRL, bridges the gap between RL and supervised learning by transforming much of the RL problem into a supervised learning task. With UDRL, the algorithm is no longer directly learning to maximize the expected return, but is instead learning a mapping between commands (e.g., desired returns) and actions. The data for training the model can be collected either offline---as in Decision Transformers~\citep{chen2021decision}---or online as the model learns to generalize to higher and higher returns. One notable extension of UDRL is GoGePo~\citep{faccio2023goal}, which extends UDRL from working in action space to working in parameter space.
However, this approach relies on a generator-evaluator (actor-critic) architecture. Here, the generator learns to produce policies that follow a given command, and the evaluator network assesses the quality of the generated policies. This introduces a considerable amount of additional complexity, as the evaluator must be jointly optimized with the generator.

We propose UDRL with Policy Generators (UDRLPG): a simpler alternative to GoGePo that sidesteps the need for an evaluator.
UDRLPG learns a single policy generator, capable of producing policies that can achieve any desired return, without relying on a generator-evaluator pair. Through hindsight learning, the model minimizes the error between policies it previously generated and new policies it produces, without a critic, thereby reducing the architectural complexity.

\section{Background}

Formally, the RL problem is often modeled as a Markov Decision Process~\citep{puterman2014markov,stratonovich1960}, or MDP, which is a tuple $(S, A, P, R, \gamma, \mu_0)$, where at each timestep $t$, the agent observes a state $s_t \in S$, chooses an action $a_t \in A$, and receives a reward $r_t = R(s_t, a_t)$. The action leads to a new state according to the transition probability $P(s_{t+1}|s_t, a_t)$. Each episode begins in an initial state $s_0$ selected with probability $\mu_0$. The policy $\pi_\theta: S \rightarrow \Delta(A)$ controls the agent, where $\theta \in \Theta$ are the policy parameters. The objective is to find $\pi_{\theta}$ that maximizes the expected return: $\pi_{\theta} = \arg\max_{\pi_\theta} J(\theta)$, where $J(\theta) = \int_{\mathcal{T}} p(\tau|\theta)R(\tau) d\tau$. Here, $p(\tau|\theta)$ is the distribution over trajectories induced by policy $\pi$ with parameters $\theta$.

Goal- and command-conditioned RL~\citep{schmidhuber1991learning,schmidhuber1993planning,schaul2015universal,andrychowicz2017hindsight} agents differ from classic RL agents as they learn to maximize a goal- or command-conditioned expected return.
UDRL and related approaches employ supervised learning to train command-conditioned RL agents by receiving command inputs that specify the desired outcome within a certain timeframe. However, in the episodic setting, it is often the case that there is no single sequence of actions or behaviors that satisfies a given command. For this reason, a unimodal maximum likelihood approach may not be able to capture the variability in the data.

UDRL takes the problem of learning to act within an environment closer to a supervised learning task, as the goal of the model now becomes learning a mapping from state and command to actions rather than learning a mapping from state-action pair to value (expected return) or maximizing return directly using policy search. The primary benefit of this formulation lies in its ability to convert a portion of the RL problem into a supervised task. This allows us to handle part of the complexity of reinforcement learning problems within the supervised learning domain, which is the main area where artificial neural networks are most successfully applied. In UDRL, the agent is trained using hindsight. It learns to predict which action it took, given the current state and command $(g, h)$, where $g$ is the actual return observed. UDRL also requires the definition of a command selection strategy. Typically, for online RL reward-maximization problems, the commands being issued should increase over time.

Building on the paradigm introduced by Fast Weight Programmers (FWPs), where one neural network learns to generate weight updates for another network, parameter-based methods in RL~\citep{sehnke_parameterexploring_2010, sehnkepgpecontrol,salimans2017evolution,mania2018simple} sample policy parameters $\theta$ from a hyperpolicy distribution $\nu_\rho(\theta)$~\citep{faccio2023goal}, transforming the RL problem from finding the parameters $\theta$ of a policy $\pi$ such that the
expected return obtained in the environment is maximized, to finding the hyperpolicy parameters $\rho$ that maximizes the expected return $J(\rho) = \int_{\Theta}\nu_\rho(\theta) \int_{\mathcal{T}} p(\tau|\theta)R(\tau) , d\tau d\theta$. The objective can be made context-dependent as $J(\rho, c)$ by conditioning the hyperpolicy on context $c$, $J(\rho, c) = \int_{\Theta}\nu_\rho(\theta | c) \int_{\mathcal{T}} p(\tau|\theta)R(\tau) \, d\tau d\theta.$

\section{Related Work}

The development of UDRLPG builds upon several foundational concepts in RL and neural networks. FWPs~\citep{schmidhuber1992learning, schmidhuber1993self}, introduced the concept of using one neural network to output weight updates for another target network---including implementations with deep and recurrent neural architectures~\citep{schmidhuber1992steps}---enabling dynamic context-dependent weights after training. This concept was later popularized under the name Hypernetworks. FWPs have found applications across various domains, including memory-based meta learning~\citep{schmidhuber1993self, miconi2018differentiable} and RL~\citep{gomez2005co}.

Parameter-based value functions~\citep{faccio2021parameterbased,harb2020policy} abstract traditional value functions by learning to estimate expected returns conditioned on policy parameters over states or state-action pairs. This enables evaluation and adjustment in parameter space, providing a mapping between policy parameters and expected returns. The concept of inverting this mapping---determining parameters given an expected return---forms a core principle of UDRLPG.
UDRLPG removes the need for the evaluator function  $V_w: \Theta \rightarrow \mathbb{R}$ that GoGePo uses to to optimize a policy generator. The generator in GoGePo learns to minimize the difference between return commands and estimated returns of generated policies through parameter-based evaluation.

Evolution Strategies demonstrate the effectiveness of directly exploring policy parameter space, showing that parameter space optimization can overcome limitations of action-space methods like sparse rewards. Policy Gradients with Parameter-Based Exploration addresses high variance in policy gradient methods by replacing policies with probability distributions over parameters, enabling trajectory sampling from single parameter samples~\citep{salimans2017evolutionstrategiesscalablealternative}.

\section{Method}

UDRLPG is a parameter-based method that directly optimizes over policy space to generate policies achieving desired returns. At its core, UDRLPG employs a FWP, in the form of a hypernetwork \( G_\rho : \mathbb{R}^{n_c} \rightarrow \Theta \) functioning as a decoder, where \( c \in \mathbb{R}^{n_c} \) represents the command (desired return) and \( \rho \) are the FWP parameters. For exploration purposes, we consider a non-deterministic FWP \( g_\rho(\theta, c) = G_\rho(c) + \epsilon \), where \( \epsilon \sim \mathcal{N}(0, \sigma^2I) \). Here, $\sigma$ is a hyperparameter that controls the extent
of the perturbation, therefore directly controlling the exploration-exploitation balance. Higher \( \sigma \) values enable broader exploration and help escape local optima, while lower values favor exploitation of known high-performing policies. The introduction of noise is vital for the learning process as it allows the algorithm to explore a wider range of potential policies, lowering the chance of getting stuck in a local optimum.

The hypernetwork $G_\rho$ is trained to minimize the error \( \mathcal{L}_G(\rho)=\mathbb{E}_{c\in D}[(G_{\rho}(c)- \theta^{*}(c))^2] \), where \( D \) represents the replay buffer and \( \theta^*(c) \) is a policy with expected return \( c \). The replay buffer is initialized with random policies to ensure diverse starting conditions, enabling effective exploration during the early stages of training. As a UDRL method, UDRLPG performs the usual hindsight learning~\citep{andrychowicz2017hindsight}, which uses past experiences as examples of successfully following specific commands. Generated policies are stored in the replay buffer together with their observed return replacing the command (desired return), rather than the command (desired return).

The training process consists of an update and a rollout stage. The first stage is the update, where the hypernetwork undergoes typical iterative gradient updates using policies sampled from the replay buffer. During the rollout phase, new policies are generated using the updated weights, and, after noise is added for exploration, the policies are evaluated using observation normalization and added to the replay buffer. To address overrepresentation of particular return ranges during training, the buffer is organized into performance-based buckets containing policies within specific return ranges. This organization decouples sampling probability from the absolute number of policies in each performance category, allowing one to follow the desired weighting strategy independently of the returns of policies in the buffer.

A full implementation of UDRLPG is publicly available at \href{https://github.com/JacopoD/udrlpg_}{\texttt{https://github.com/JacopoD/udrlpg\_}}

\section{Results and Discussion}

We compare UDRLPG to two baseline algorithms: GoGePo and DDPG in the \texttt{InvertedPendulum-v4}, \texttt{Swimmer-v4}, and \texttt{Hopper-v4} environments from the OpenAI gym suite~\citep{brockman2016openai}. For each environment, we report the mean return and the variance of the last few returns.
We also analyze the model's ability to produce policies across the return spectrum.

\begin{figure}
    \centering
    \begin{minipage}{0.1\textwidth}
        \centering
        \includegraphics[width=\linewidth]{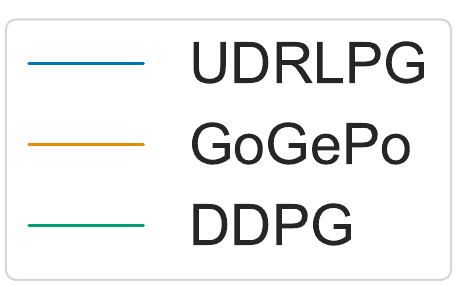}
    \end{minipage}\hfill
    \begin{minipage}{0.29\textwidth}
        \centering
        \includegraphics[width=\linewidth]{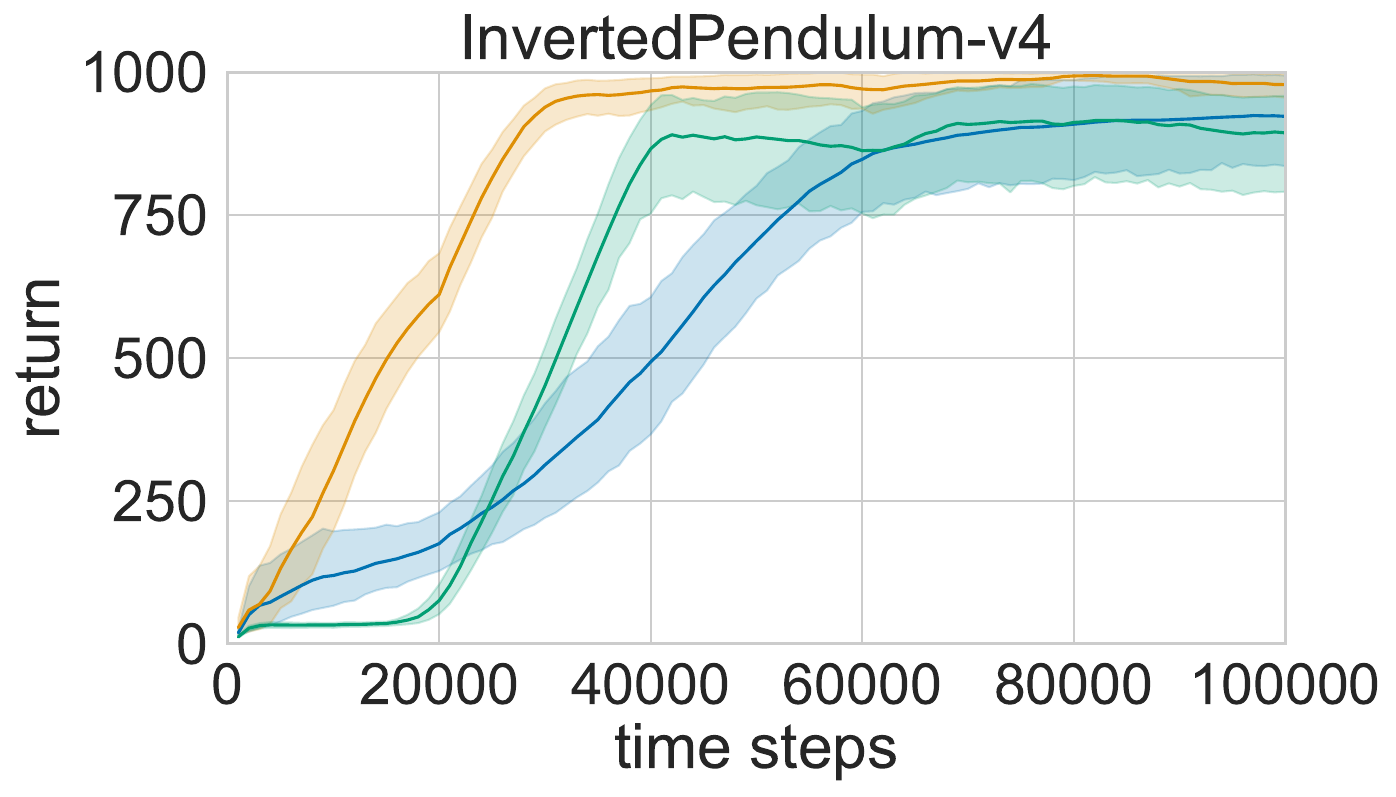}
    \end{minipage}\hfill
    \begin{minipage}{0.29\textwidth}
        \centering
        \includegraphics[width=\linewidth]{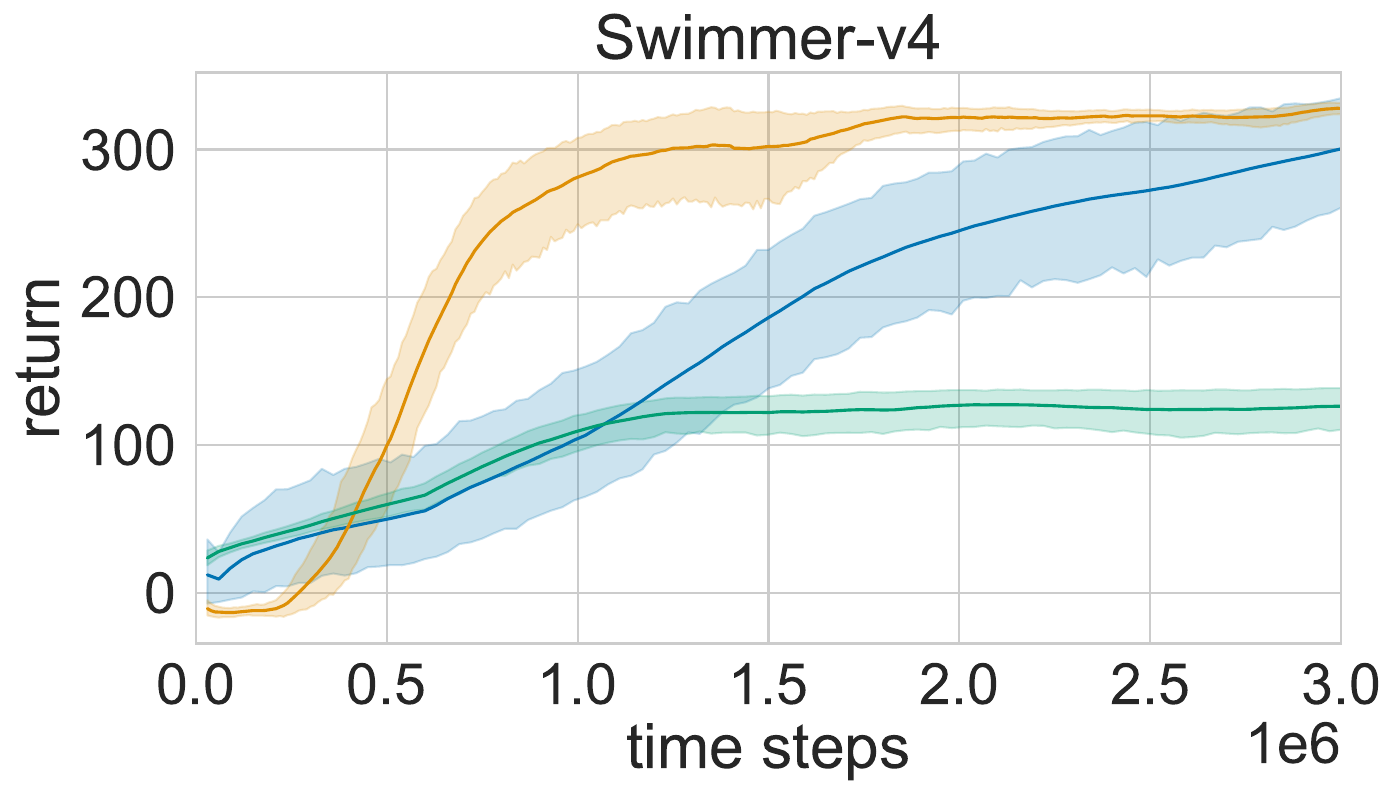}
    \end{minipage}\hfill
    \begin{minipage}{0.29\textwidth}
        \centering
        \includegraphics[width=\linewidth]{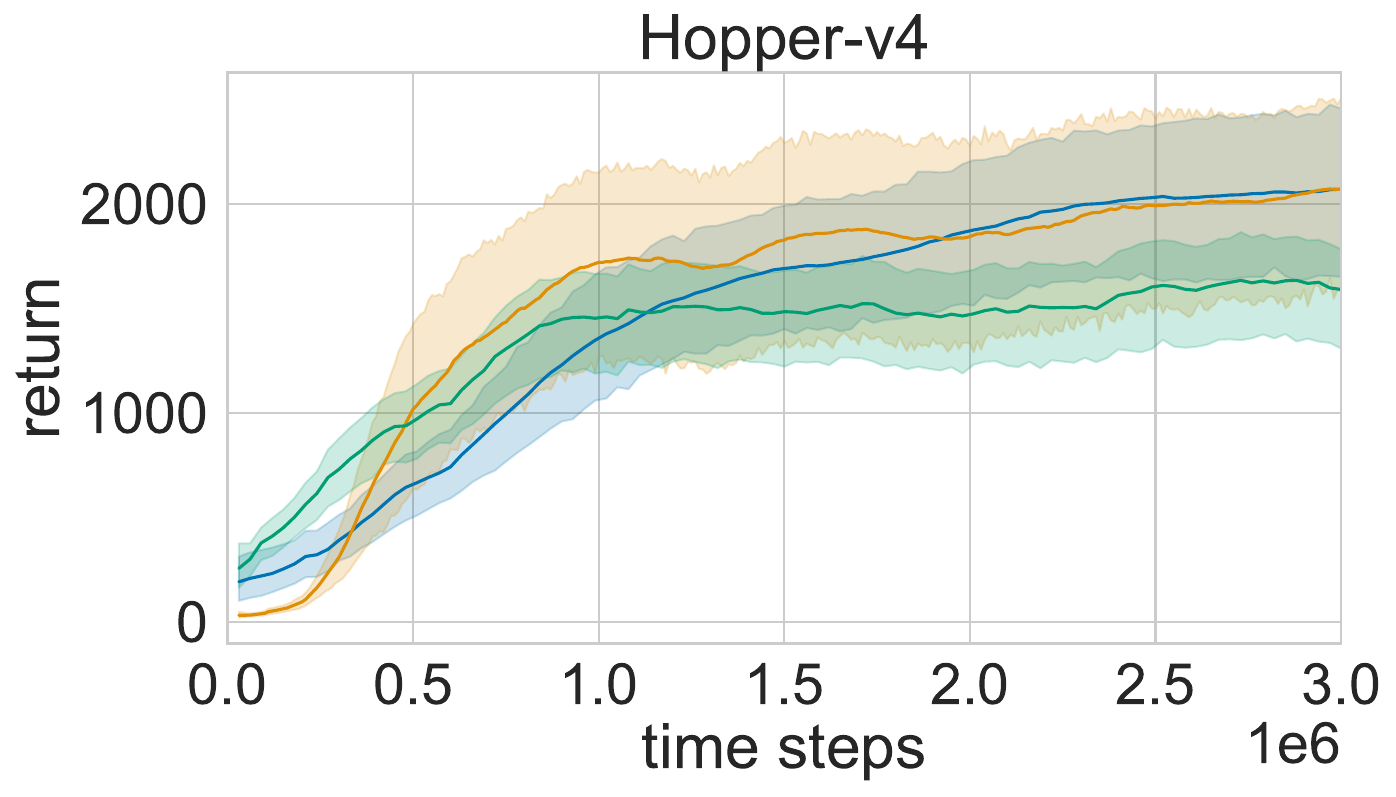}
    \end{minipage}
    \caption{Performance of policies from UDRLPG, GoGePo, and DDPG during training in all environments. Lines show mean return and $95\%$ bootstrapped confidence intervals from $20$ independent runs.}%
    \label{fig:returns}
\end{figure}

\begin{figure}
    \centering
    \begin{minipage}{0.17\textwidth}
        \centering
        \includegraphics[width=\linewidth]{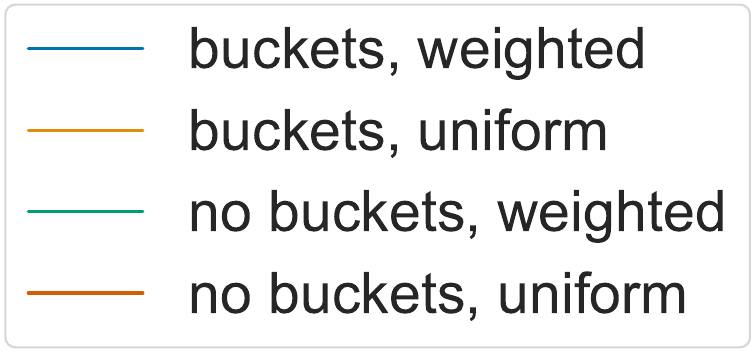}
    \end{minipage}\hfill
    \begin{minipage}{0.27\textwidth}
        \centering
        \includegraphics[width=\linewidth]{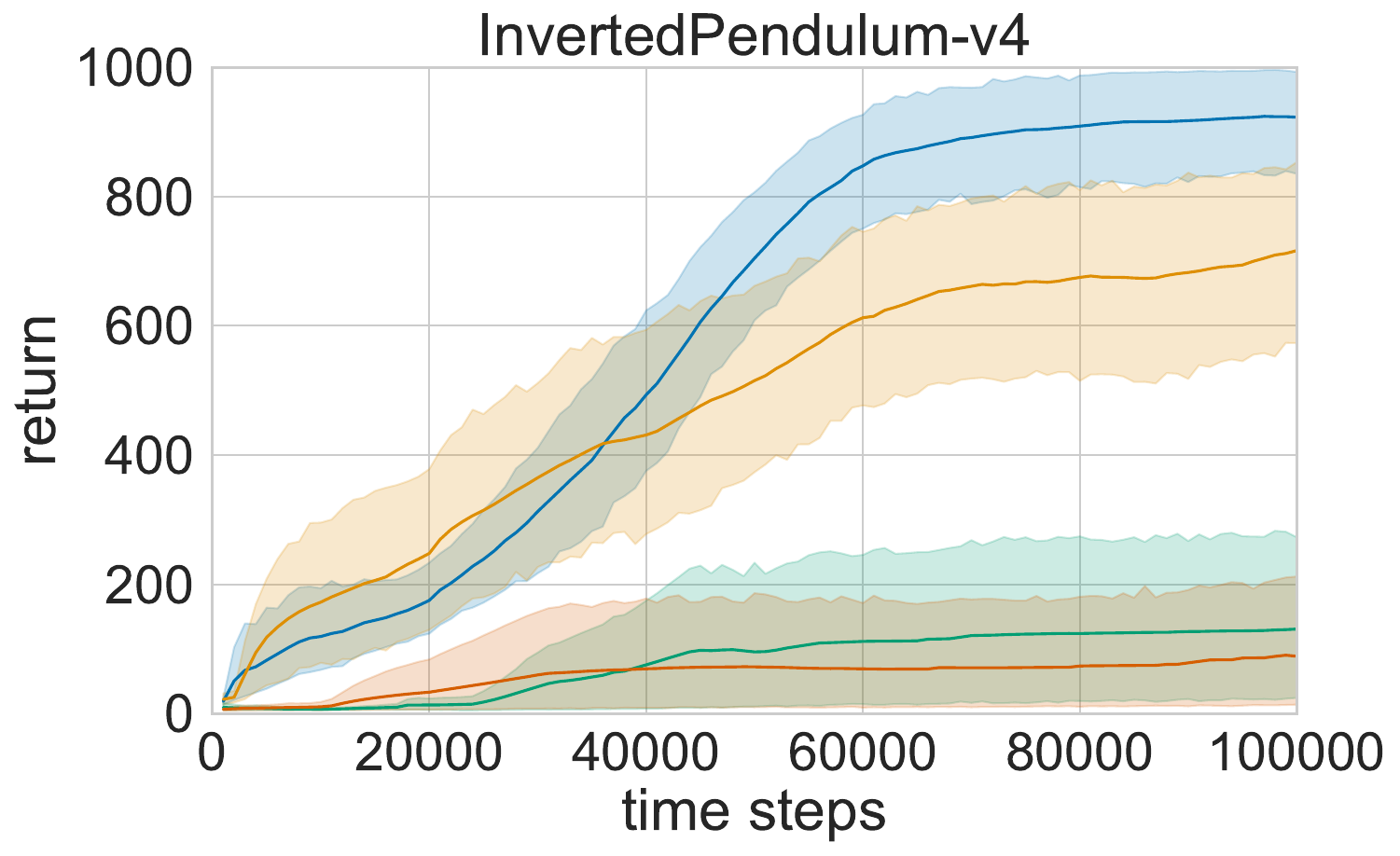}
    \end{minipage}\hfill
    \begin{minipage}{0.27\textwidth}
        \centering
        \includegraphics[width=\linewidth]{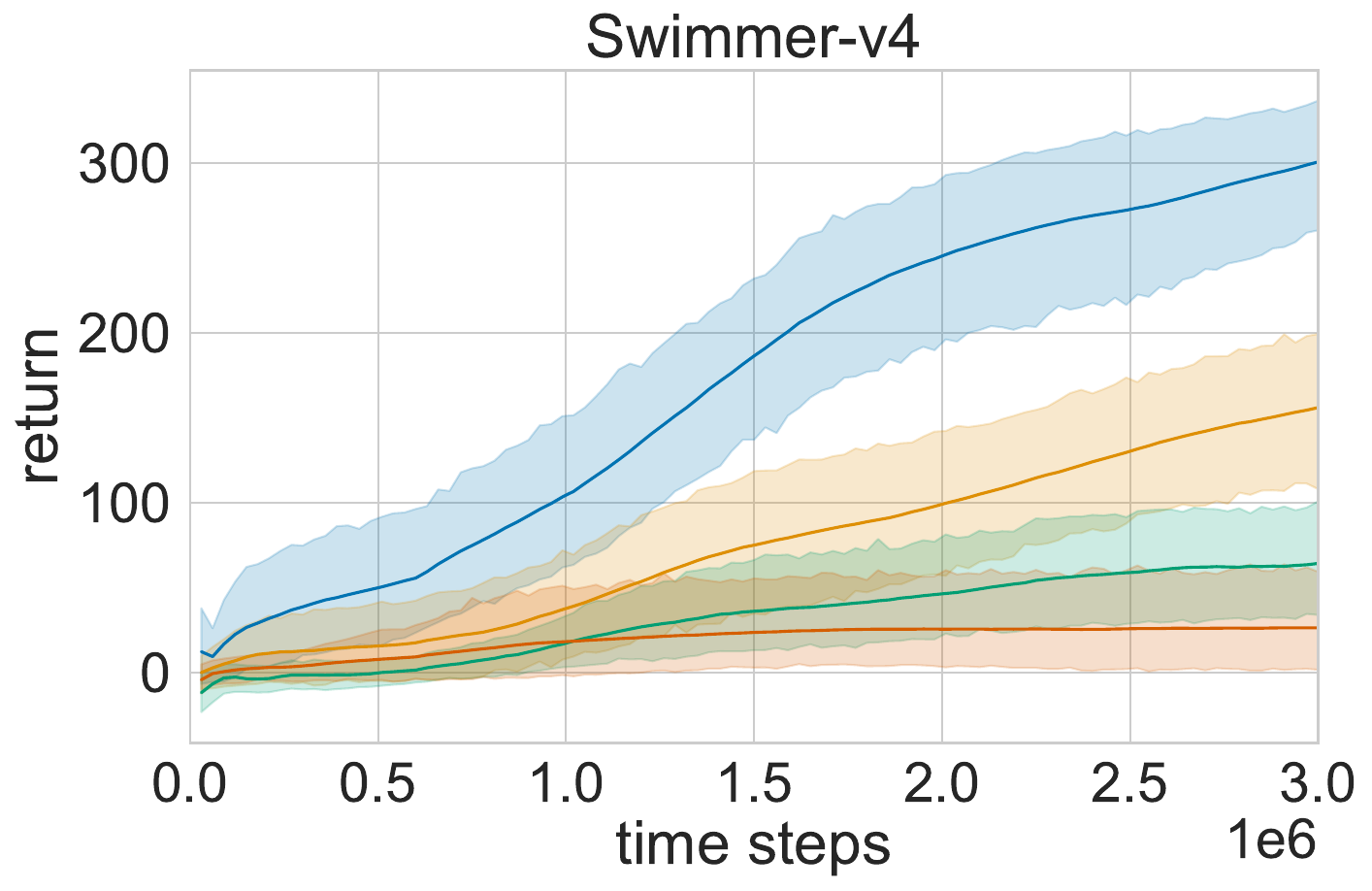}
    \end{minipage}\hfill
    \begin{minipage}{0.27\textwidth}
        \centering
        \includegraphics[width=\linewidth]{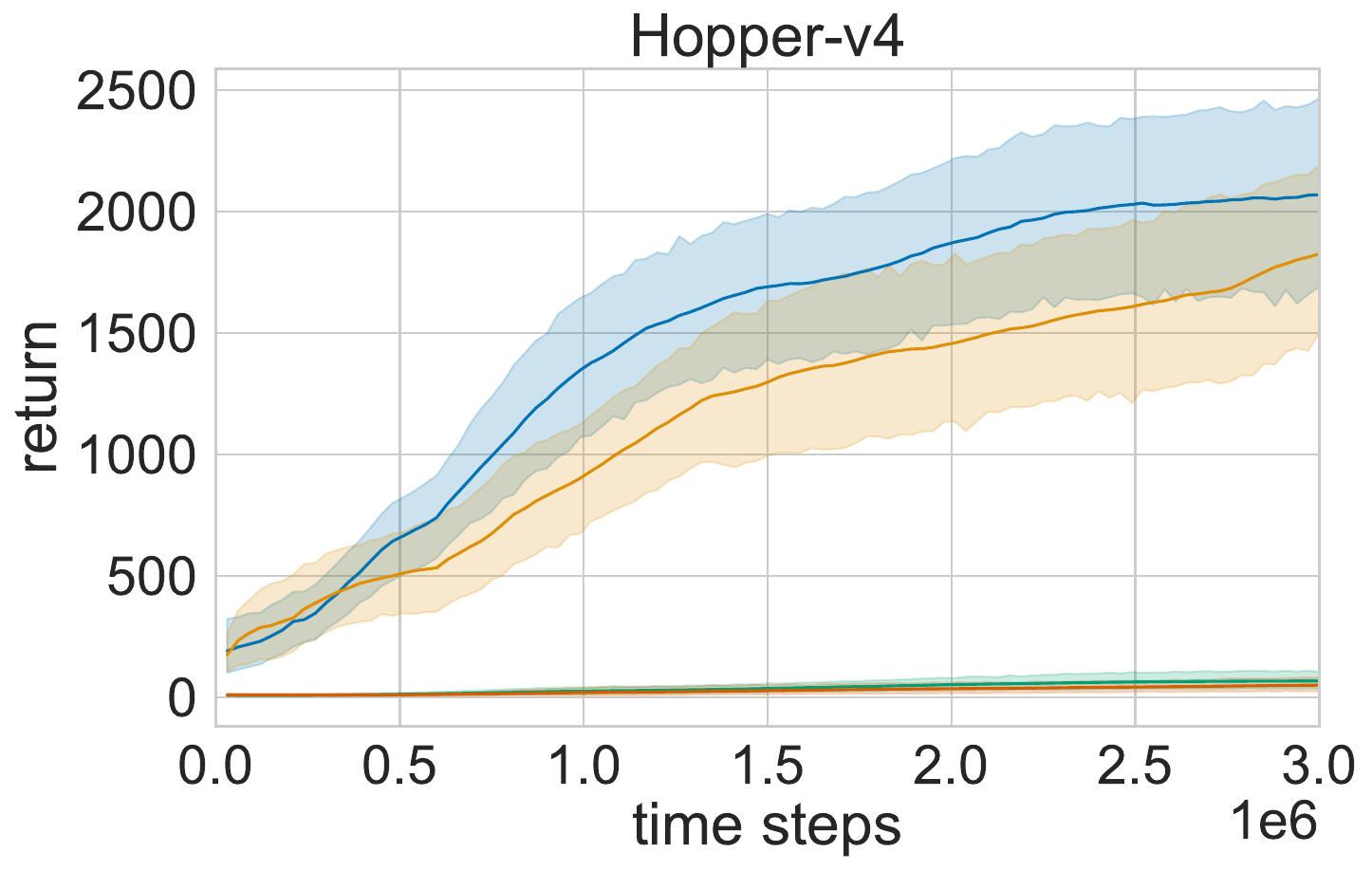}
    \end{minipage}
    \caption{Performance comparison of UDRLPG policies across test all environments using four buffer strategies. The proposed strategy is buckets and weighted sampling, in blue. Curves show mean returns with $95\%$ bootstrapped confidence intervals from $20$ runs.}
    \label{fig:ablation}
\end{figure}

\begin{figure}
    \centering
    \begin{minipage}{0.29\textwidth}
        \centering
        \includegraphics[width=\linewidth]{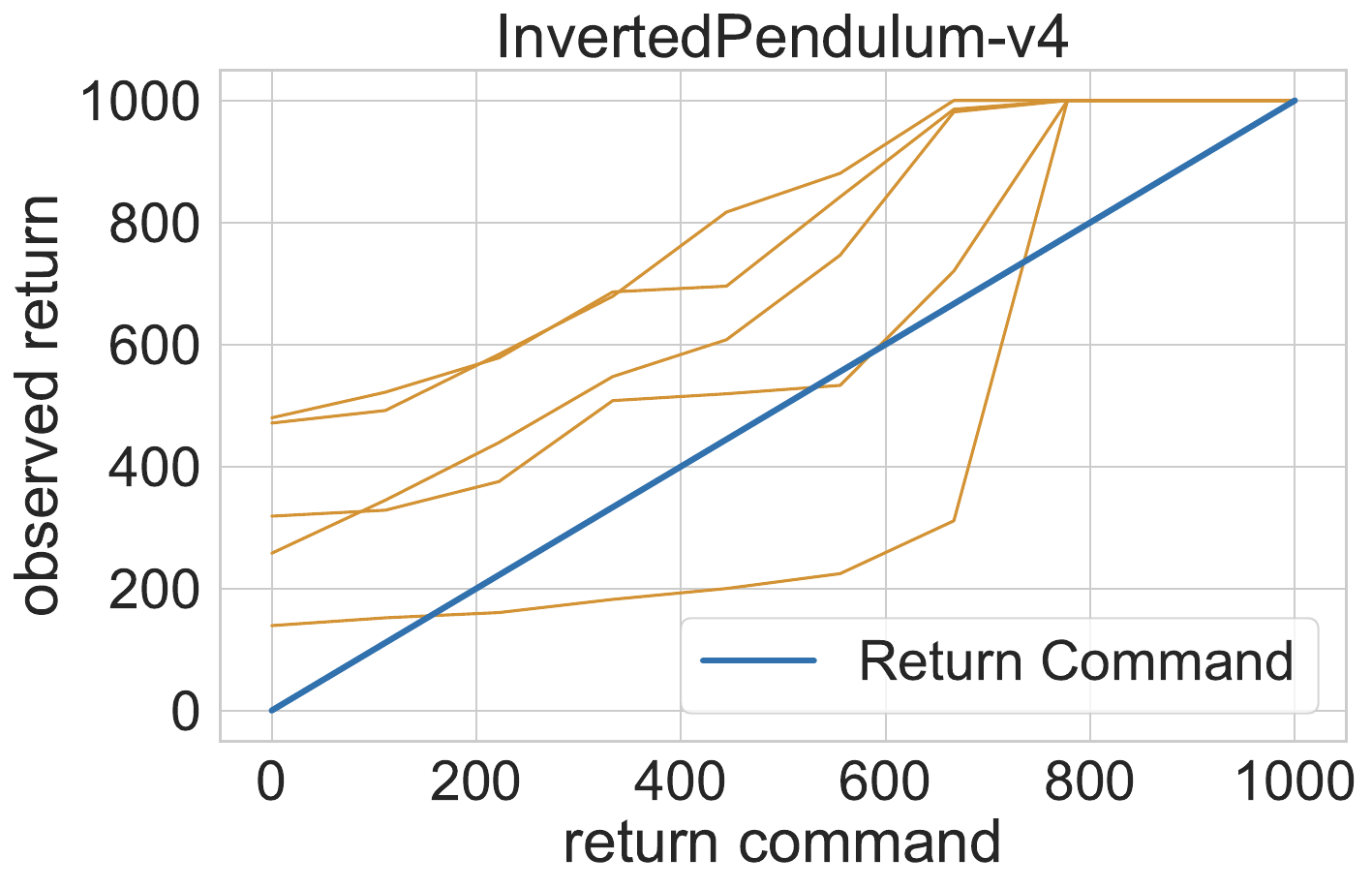}
    \end{minipage}\hfill
    \begin{minipage}{0.29\textwidth}
        \centering
        \includegraphics[width=\linewidth]{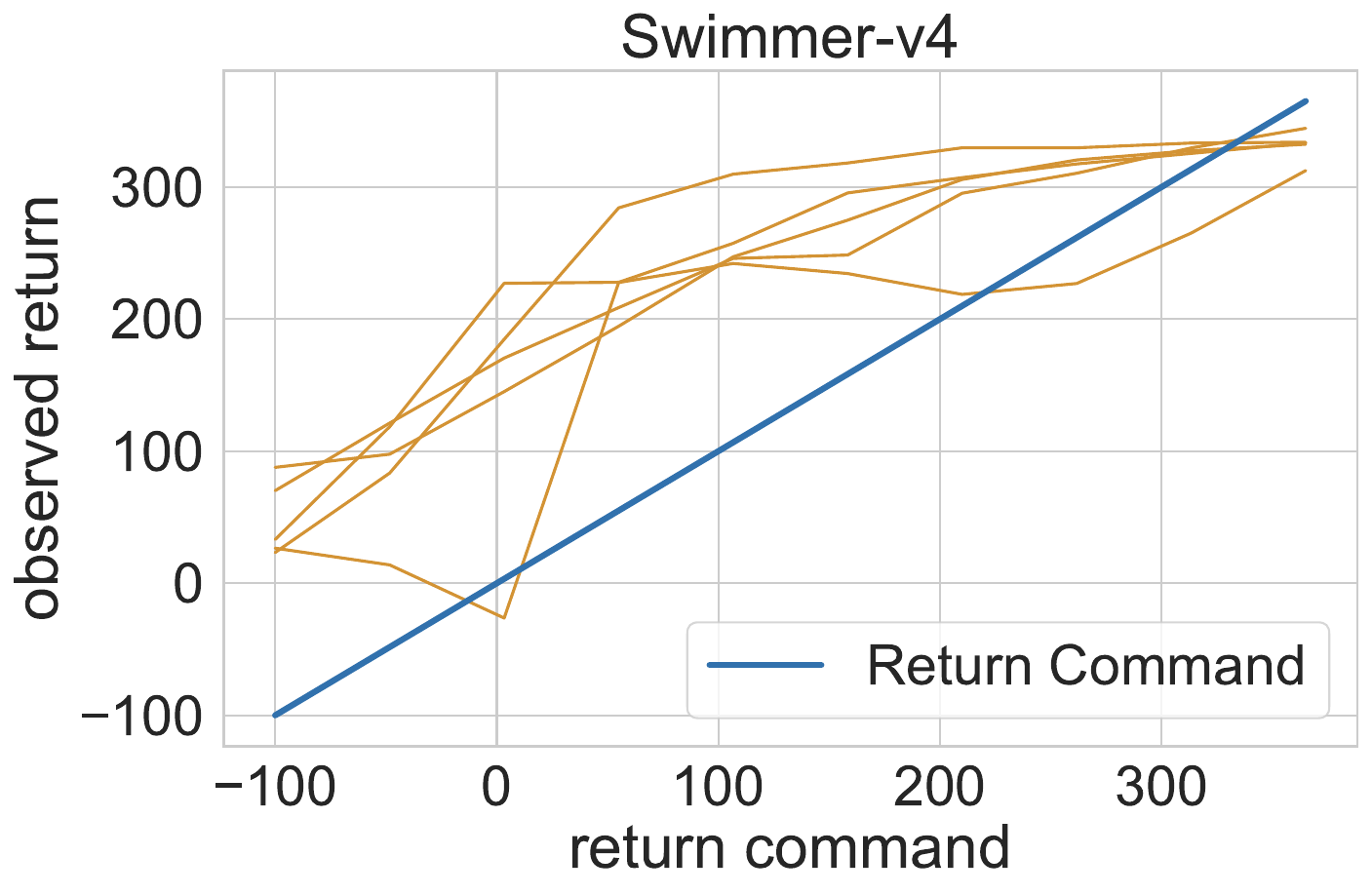}
    \end{minipage}\hfill
    \begin{minipage}{0.29\textwidth}
        \centering
        \includegraphics[width=\linewidth]{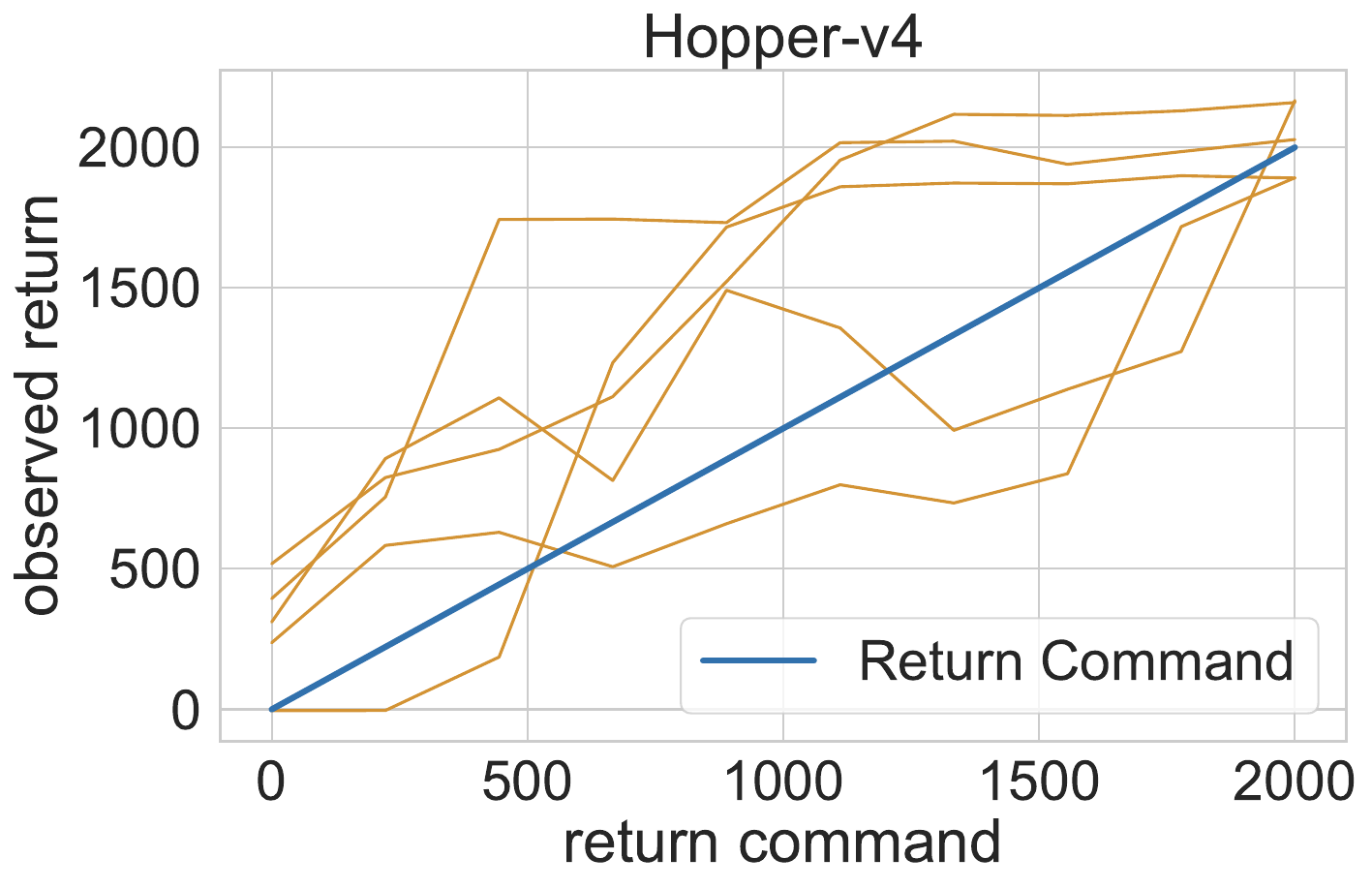}
    \end{minipage}
    \caption{Mean returns over $10$ episodes of policies from the generator as a function of the given command. Results are averaged over $5$ independent runs.}
    \label{fig:identity}
\end{figure}

Results in Figure~\ref{fig:returns}, show competitive performance against both baselines. In InvertedPendulum, UDRLPG converges to the same value but slower than both baselines.
While it achieves the maximum possible reward of $1000$, it exhibits higher variance in final returns compared to GoGePo, indicating less stability across runs. For Swimmer, UDRLPG reaches a mean final return of $300$, underperforming against GoGePo which reaches $320$. The method shows a steady improvement throughout training with no signs of plateauing, with higher variance in final performance compared to GoGePo. In Hopper, the hardest environment tested, UDRLPG matches the performance of GoGePo with a mean final return of $2070$. UDRLPG appears to explore the parameter space more extensively than the baselines, resulting in higher return bounds. As shown in Figure~\ref{fig:identity}, UDRLPG can produce policies across the return spectrum, resulting in strong identity curves, suggesting robust generalization over commands. Performance-based buckets and a fine-tuned weighting strategy for the replay buffer were crucial for stable training. This approach reduces learning stagnation and ensures a balanced representation of high and low return policies during training, leading to more consistent convergence toward higher return policies. Our ablation experiments, shown in Figure~\ref{fig:ablation}, provide empirical evidence for this claim.

UDRLPG inherits from UDRL potential challenges deriving from multimodality. The newly generated policy $\theta_{new}$, obtained by conditioning the generator on a command $c$ may be significantly different from the policies $\theta$ in the buffer, where $R(\theta)=c$. If $|D(c)|\ge 2$, the generator will try to fit $\theta_{mean}$: the mean of the policies. However, there is no guarantee that $R(\theta_{mean}) = R(\theta)$. In this case, the learning process may start degrading. This issue does not arise frequently while training UDRLPG. One possible explanation is that, early on during training, the hypernetwork develops an understanding of the underlying structure or distribution of the policy weights based on the first policies in the buffer. Weight initialization biases the hypernetwork towards a specific arrangement of the neural network's weights~\citep{chen1993geometry}. This arrangement includes the order of the neurons within hidden layers (neuron permutation) and the magnitudes of the weights (weight scaling). The bias induced by the initialization constrains the search space in a part of the solution space where all weights follow the same configuration. The hypernetwork learns changes around the configuration, reducing the likelihood of generating policies with similar returns that are very different in weight space.

\begin{figure}
    \centering
    \begin{minipage}{0.3\textwidth}
        \centering
        \includegraphics[width=\linewidth]{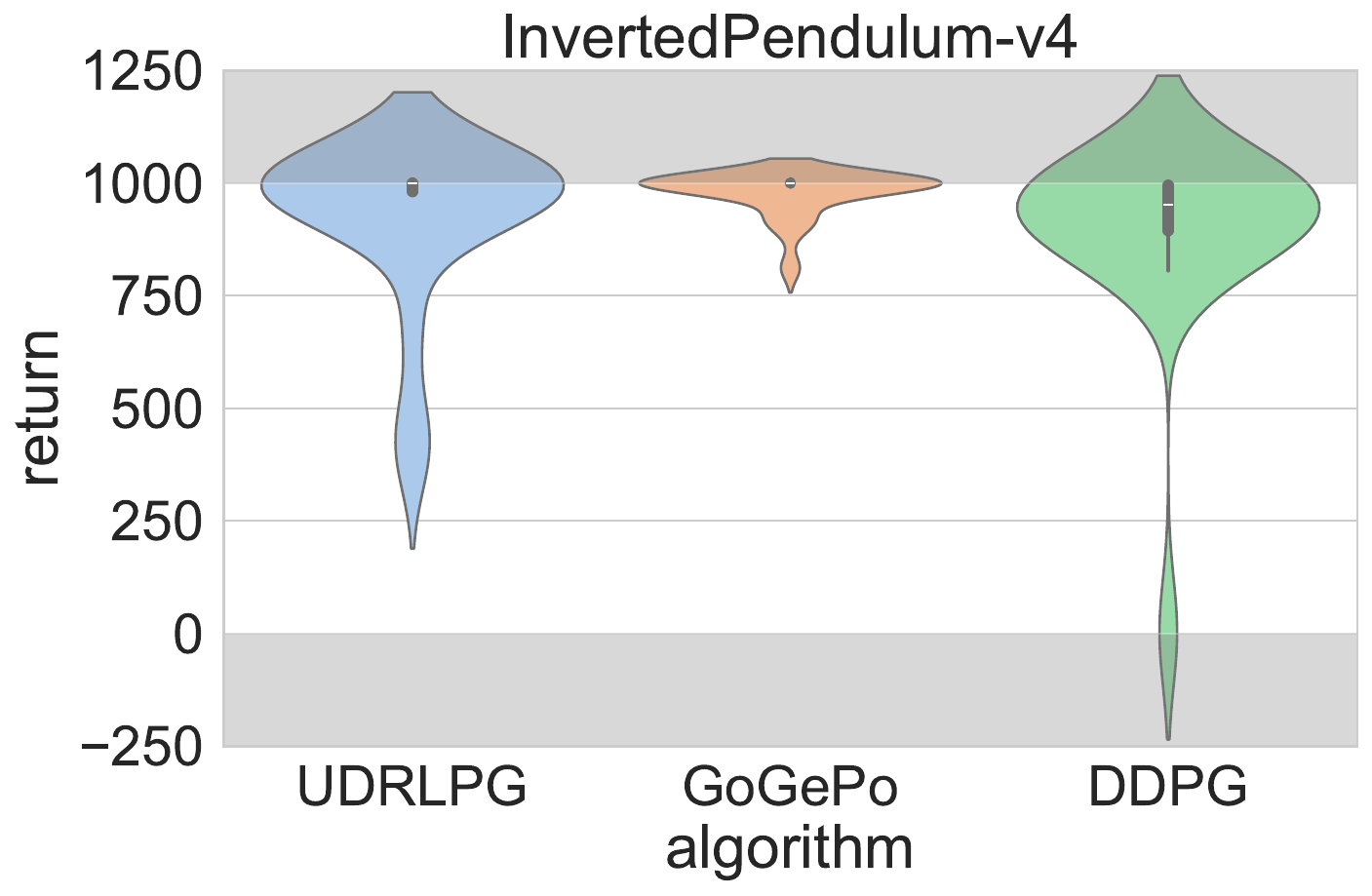}
    \end{minipage}\hfill
    \begin{minipage}{0.3\textwidth}
        \centering
        \includegraphics[width=\linewidth]{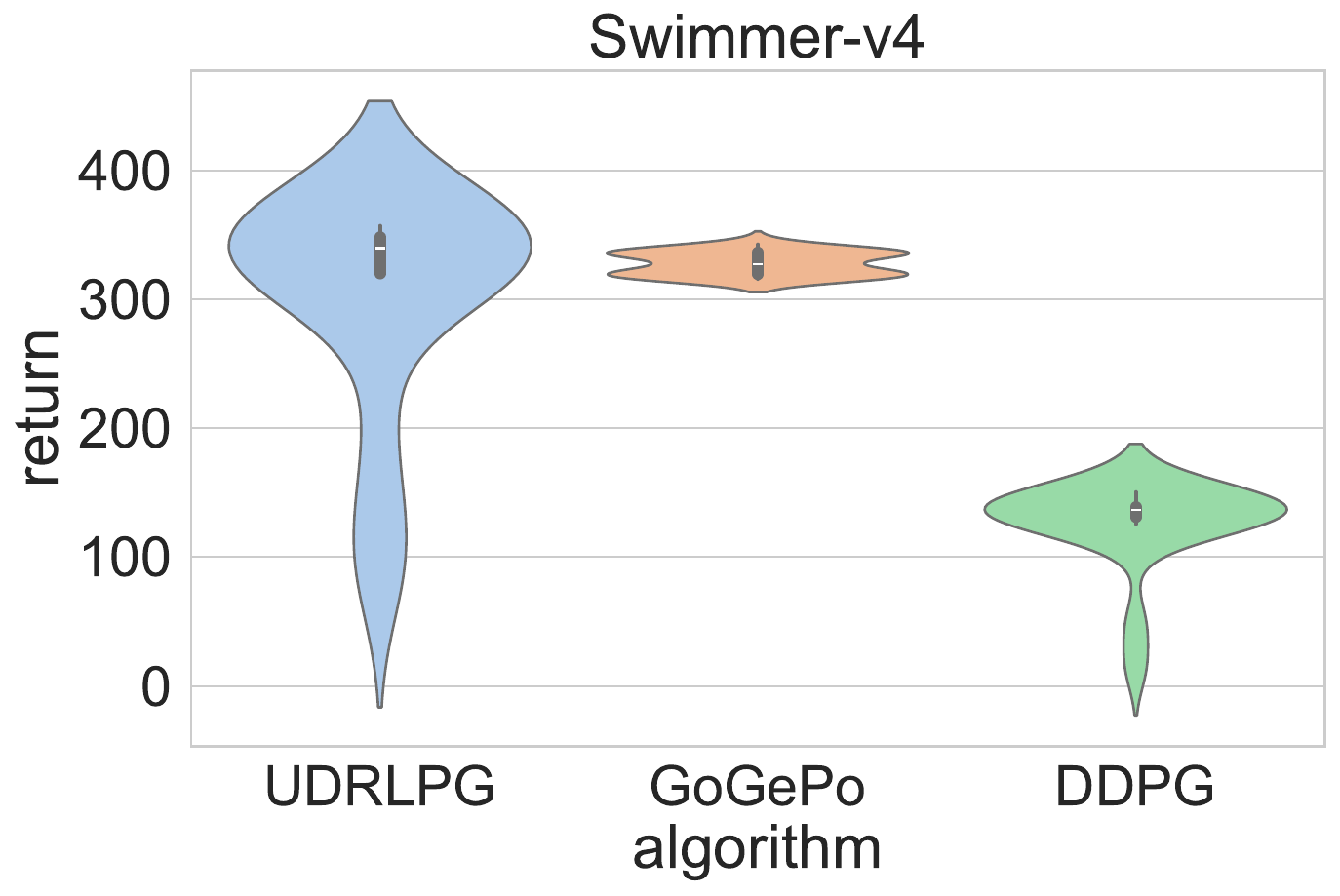}
    \end{minipage}\hfill
    \begin{minipage}{0.3\textwidth}
        \centering
        \includegraphics[width=\linewidth]{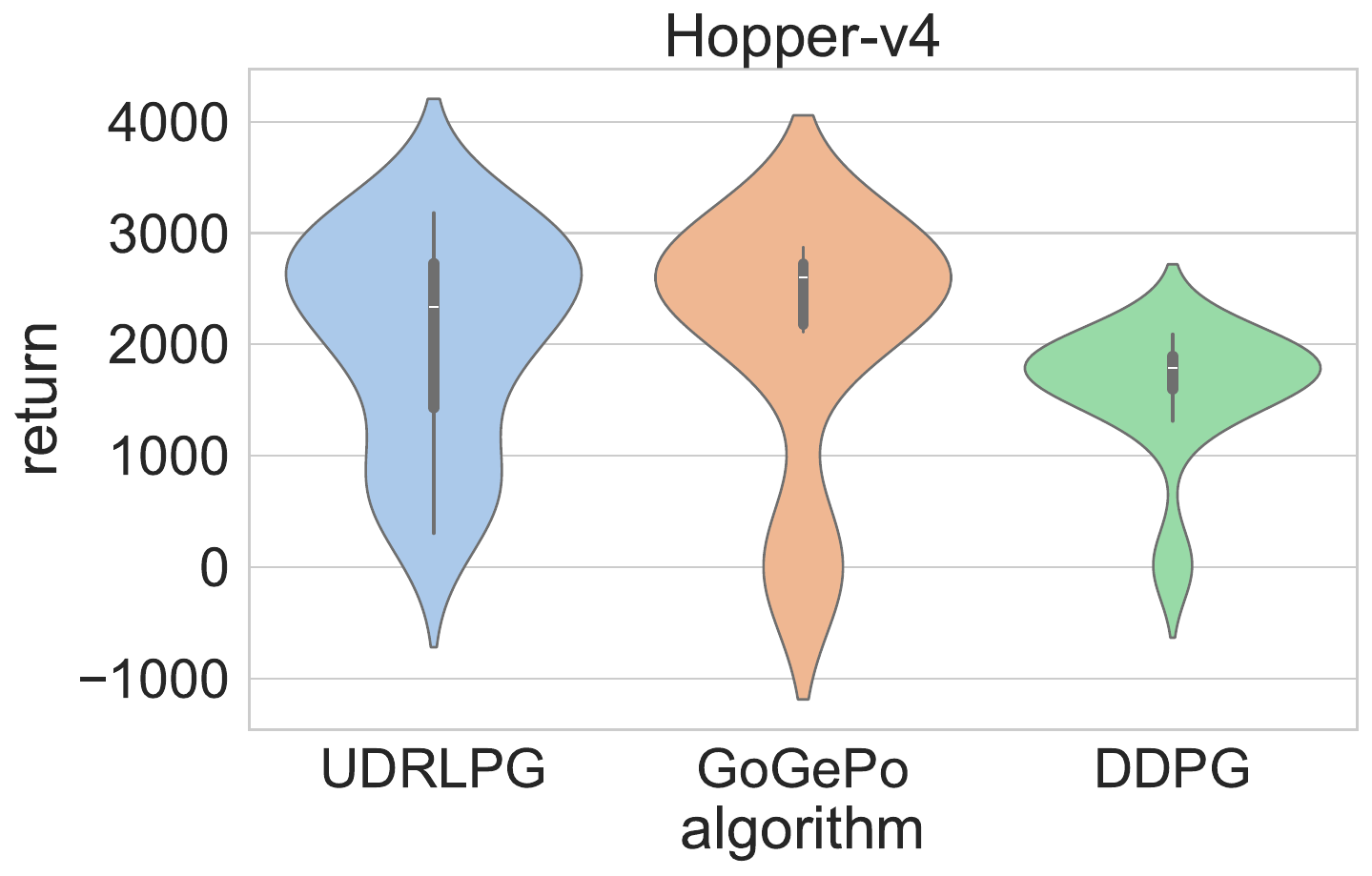}
    \end{minipage}
    \caption{Variance of final returns. Lines show mean and $95\%$ confidence intervals from $20$ evaluation runs.}%
    \label{fig:violin}
\end{figure}

\section{Conclusion}

This work introduced UDRLPG, an approach to RL focused on generating policy parameters conditioned on return commands.
Compared with existing methods, UDRLPG removes the need for a separate evaluator in the architecture, thus simplifying the overall structure.
Empirical results show that UDRLPG generalizes effectively across commands, is competitive with existing methods, and is able to
explore the parameter space more extensively than some competing methods, resulting in higher return bounds.
Additionally, we note that the hypernetwork's initialization bias confines the search to a specific region of the solution space where weights share a common configuration, effectively circumventing the challenge of multimodality.
We identify some limitations of UDRLPG here.
In some environments, convergence is slower and the variance in final returns across runs is higher than that of GoGePo. Compared to GoGePo, UDRLPG simplifies the learning process and provides further insight into goal-conditioned policy generation.

\printbibliography

\end{document}